\documentclass[12pt]{article}
\usepackage[utf8]{inputenc}
\usepackage{yfonts}
\usepackage[margin=0.8in]{geometry} 
 
\usepackage{stmaryrd}
\usepackage{graphicx}
\usepackage{tikz,pgfplots,ulem}
\pgfplotsset{compat=1.17}
\usetikzlibrary{arrows.meta}
\usepackage{varwidth}
\usepackage[parfill]{parskip}
\usepackage{breakcites}
\graphicspath{ {images/} }

\usepackage{mathrsfs} 
\usepackage{algorithmic}
\usepackage{color, colortbl}
\usepackage{mathtools}
\usepackage{setspace}

\usepackage{subcaption}

\usepackage{bbding}
\usepackage{pifont}
\usepackage{wasysym}
\usepackage[colorlinks = true,
            linkcolor = blue,
            urlcolor=blue,
            citecolor = blue]
            {hyperref}
\usepackage{enumitem}
\usepackage{float}
\usepackage{multirow}
\usepackage{authblk}
\usepackage{siunitx}
\usepackage{gensymb}
\usepackage{amsmath,amsthm,amssymb,amsfonts,bm}

\usepackage{lineno}
\usepackage{xcolor}

\usepackage[ruled,linesnumbered]{algorithm2e} 

\newcommand{\nosemic}{\renewcommand{\@endalgocfline}{\relax}}% Drop semi-colon ;
\newcommand{\dosemic}{\renewcommand{\@endalgocfline}{\algocf@endline}}% Reinstate semi-colon ;
\let\oldnl\nl% Store \nl in \oldnl
\newcommand{\nonl}{\renewcommand{\nl}{\let\nl\oldnl}}% Remove line number for one line
\usepackage{tabularx}
\newcolumntype{Y}{>{\centering\arraybackslash}X}
\usepackage{array}

\usepackage{tikz}
\usetikzlibrary{shapes,backgrounds,arrows.meta,positioning,backgrounds,fit,shadows}
\tikzset{
 baseLine/.style={draw, fill=blue!20, text width=30, minimum width=70, text centered, minimum height=25}, %,drop shadow
 back group1/.style={fill=yellow!20, rounded corners, draw=black!50, dashed, inner xsep=15, inner ysep=20},
 back group2/.style={fill=yellow!20,rounded corners, draw=black!50, dashed, inner xsep=45, inner ysep=27.5},
 line/.style={draw, thick, color=black!50, -LaTeX},
}
\usetikzlibrary{calc}
\definecolor{ao-E}{rgb}{0.0, 0.5, 0.0} %LM
\definecolor{airforceblue}{rgb}{0.36, 0.54, 0.66} %LM
\definecolor{cadmiumgreen}{rgb}{0.0, 0.42, 0.24} %LM
\definecolor{LM}{rgb}{0.13, 0.5, 0.85} %LM
\definecolor{mygray}{RGB}{150,150,150}
\definecolor{lightblue}{RGB}{0,176,240}
\definecolor{lightgray}{RGB}{200,200,200}
\usepackage[most]{tcolorbox}
\definecolor{BulletsColor}{rgb}{0, 0, 0.9}
\newlist{myBullets}{itemize}{1}

\setlist[myBullets]{
  label=\textcolor{black}{\textbullet},
  leftmargin=*,
  topsep=0ex,
  partopsep=0ex,
  parsep=0ex,
  itemsep=0ex,
  before={\color{black}}
}

\definecolor{Gray}{gray}{0.9}
\usepackage{array}

\usepackage{algorithmic}

\usepackage{graphicx}
\usepackage{tikz,pgfplots}
\usepgfplotslibrary{groupplots}
\usepackage{adjustbox}

\usepackage{hyperref}
\hypersetup{
    colorlinks = true,
    linkbordercolor = {white},
    citecolor = black,
    urlcolor  = black,
    linkcolor = black,
}
\usepackage[utf8]{inputenc} % allow utf-8 input
\usepackage[T1]{fontenc}    % use 8-bit T1 fonts
\usepackage{hyperref}       % hyperlinks
\usepackage{url}            % simple URL typesetting
\usepackage{booktabs}       % professional-quality tables
\usepackage{amsfonts}       % blackboard math symbols
\usepackage{nicefrac}       % compact symbols for 1/2, etc.
\usepackage{xcolor}
\usepackage{microtype}      % microtypography
\usepackage{amsmath}

\usepackage{graphicx}
\usepackage{tabularx}
\newcolumntype{Y}{>{\centering\arraybackslash}X}
\newcolumntype{B}{X}
\newcolumntype{S}{>{\hsize=.478\hsize}X}
\newcolumntype{T}{>{\hsize=.29\hsize}X}
\title{Graph Convolutional Network-based Feature Selection for High-dimensional and Low-sample Size Data}

\date{}
\author[1]{Can Chen}
\author[1]{Scott T. Weiss}
\author[1,2,$\dagger$]{Yang-Yu Liu}

\affil[1]{\small{Channing Division of Network Medicine, Department of Medicine, Brigham and Women's Hospital and Harvard Medical School, Boston, MA 02115, USA}}
\affil[2]{Center for Artificial Intelligence and Modeling, the Carl R. Woese Institute of Genomic Biology, University of Illinois at Urbana-Champaign, Champaign, IL 61820, USA}
\affil[*]{\small To whom correspondence should be addressed (yyl@channing.harvard.edu).}

\begin{document}

\maketitle

% \begin{affiliations}
%  \item Channing Division of Network Medicine, Department of Medicine, Brigham and Women's Hospital and Harvard Medical School, Boston, MA 02115, USA
%  \item Center for Artificial Intelligence and Modeling, Carl R. Woese Institute of Genomic Biology, University of Illinois at Urbana-Champaign, Champaign, IL 61820, USA
%  \item[* ] To whom correspondence should be addressed (yyl@channing.harvard.edu).
% \end{affiliations}

\begin{abstract}
Feature selection is a powerful dimension reduction technique which selects a subset of relevant features for model construction. Numerous feature selection methods have been proposed, but most of them fail under the high-dimensional and low-sample size (HDLSS) setting due to the challenge of overfitting. In this paper, we present a deep learning-based method -- GRAph Convolutional nEtwork feature Selector (GRACES) -- to select important features for HDLSS data. We demonstrate empirical evidence that GRACES outperforms other feature selection methods on both synthetic and real-world datasets. 
\end{abstract}

% \large

\section{Introduction}
Many biological data representations are naturally high-dimensional and low-sample size (HDLSS)~\cite{uffelmann2021genome,10002015global,alipanahi2015predicting,leung2003fundamentals,berrar2003practical}. RNA sequencing (RNA-Seq) is a next-generation sequencing technique to reveal the presence and quantity of RNA in a biological sample at a given moment \cite{kukurba2015rna}. RNA-Seq datasets often contain a huge amount of features (e.g., $\geq 10^5$), while the number of samples is very small (e.g., $\leq 10^3$). Analyzing RNA-Seq data is crucial for various disciplines in biomedical sciences, such as disease diagnosis and drug development~\cite{leung2003fundamentals,berrar2003practical}. However, many machine learning tasks such as feature selection likely fail on such data due to the challenge of overfitting. 

A useful technique in dealing with high-dimensional data is feature selection, which aims to select an optimal subset of features. Although the selection of an optimal subset of features is an NP-hard problem~\cite{chen1997minimum}, various compromised feature selection methods have been proposed. While feature selection methods are often grouped into filtering, wrapped, and embedded methods \cite{stanczyk2015feature},  in this work, we  classify them into five categories   -- statistics-based~\cite{bommert2020benchmark, golugula2011evaluating}, Lasso-based~\cite{tibshirani1996regression, yamada2014high}, decision tree-based~\cite{bommert2020benchmark, xu2014gradient}, deep learning-based~\cite{li2016deep,liu2017deep}, and greedy methods~\cite{aha1995comparative}, based on their learning schemes, see details in Section \ref{sec:2}. Note that  most of the methods address the curse of dimensionality under the blessing of large-sample size~\cite{liu2017deep}. Only a few of them can handle HDLSS data. The state-of-the-art feature selection methods for  HDLSS data are Hilbert-Schmidt
independence criterion (HSIC) Lasso~\cite{yamada2014high, yamada2018ultra} and deep neural pursuit (DNP)~\cite{liu2017deep}.

In this paper, we propose a graph neural network-based feature selection method -- GRAph Convolutional nEtwork feature Selector (GRACES)  -- to extract features by exploiting the latent relations between samples for HDLSS data. Inspired by DNP, GRACES is a deep learning-based method that iteratively finds a set of optimal features. GRACES utilizes various overfitting-reducing techniques, including multiple dropouts, introduction of Gaussian noises, and F-correction, to ensure the robustness of feature selection. We demonstrate that GRACES outperforms HSIC Lasso and DNP (and other baseline methods) on both synthetic and real-world datasets. 

The paper is organized into five sections. We perform a thorough literature review on feature selection (including traditional and HDLSS feature selection methods) in Section \ref{sec:2}. The main architecture of GRACES is presented in Section \ref{sec:3}. We evaluate the performance of GRACES along with several representative methods on both synthetic and real-world datasets in Section \ref{sec:4}. Finally, we discuss the computational costs of the methods and conclude  with future research directions in Section \ref{sec:5}.

\section{Related Work}\label{sec:2}
Univariate statistical tests have been widely applied for feature selection~\cite{bommert2020benchmark, golugula2011evaluating}. The computational advantage allows them to perform feature selection on extremely high-dimensional data. The ANOVA (analysis of variance) F-test~\cite{st1989analysis} is one of the most commonly used statistical methods for feature selection. The value of the F-statistic is used as a ranking score for each feature, where the higher the F-statistic, the more important is the corresponding feature~\cite{bommert2020benchmark}. Other classical statistical methods, including the student's t-test~\cite{owen1965power}, the Pearson correlation  test~\cite{meng1992comparing}, the Chi-squared test~\cite{plackett1983karl}, the Kolmogorov-Smirnov test~\cite{daniel1990kolmogorov}, the Wilks' lambda test~\cite{el2007feature}, and the Wilcoxon signed-rank test~\cite{wilcoxon1992individual}, can be applied for feature selection in a similar manner. Empirically, the ANOVA F-test is able to achieve a relatively good performance in feature selection on some HDLSS data with very low computational costs.

L1-regularization, also known as the least absolute shrinkage and selection operator (Lasso), has a powerful built-in feature selection capability for HDLSS data~\cite{tibshirani1996regression}. Lasso assumes linear dependency between input features and outputs, penalizing on the $l_1$-norm of feature weights. Lasso produces a sparse solution with which the weights of irrelevant features are zero. Yet, Lasso fails to capture nonlinear dependency. Therefore, kernel-based Lasso such as HSIC Lasso~\cite{yamada2014high, yamada2018ultra} has been developed for handling nonlinear feature selection on HDLSS data. HSIC Lasso utilizes the empirical HSIC~\cite{gretton2005measuring} to  find non-redundant features with strong dependence on outputs. HSIC Lasso outperforms other similar methods, including feature vector machine~\cite{li2005lasso},  minimum redundancy maximum relevance~\cite{peng2005feature}, sparse additive model~\cite{ravikumar2009sparse}, quadratic programming feature selection~\cite{rodriguez2010quadratic}, and centered kernel target alignment~\cite{cortes2012algorithms}. Additionally, the $l_1$-regularizer in Lasso can be compatibly incorporated to different classifiers such as logistic regression (LR Lasso) for feature selection~\cite{meier2008group}.

Decision tree-based methods are also popular for feature selection, which can model nonlinear input-output relations~\cite{bommert2020benchmark}. As an ensemble of decision trees, random forests (RF)~\cite{breiman2001random} calculate the importance of a feature based on its ability to increase the pureness of the leaf in each tree. A higher increment in leaves' purity indicates higher importance of the feature. In addition, gradient boosted feature selection (GBFS) selects features by penalizing the usage of features that are not used in the construction of each tree~\cite{xu2014gradient}. However, decision tree-based feature selection methods such as RF and GBFS require large-sample size for training. Hence, these methods often do not perform well under the HDLSS setting.

Numerous deep learning-based methods have been proposed for feature selection \cite{li2016deep, shrikumar2017learning, lu2018deeppink, gui2019afs, chen2017kernel, borisov2019cancelout, mirzaei2020deep, wojtas2020feature}. Like decision tree-based methods, deep neural networks also require a large number of samples for training, so these methods often fail on HDLSS data.
Nevertheless, there are several deep learning-based feature selection methods which are  designed specifically for HDLSS data \cite{liu2017deep, li2022deep}. DNP  learns features by using a multilayer perceptron (MLP) and incrementally adds them through multiple dropout technique in a nonlinear way~\cite{liu2017deep}. DNP overcomes the issue of overfitting resulting from low-sample size and outperforms other methods such as LR Lasso, HSIC Lasso, and GBFS on  HDLSS data. An alternative to DNP with replacing the MLP by a recurrent neural network is mentioned in~\cite{chowdhury2019recurrent}. Yet, DNP only uses MLP to generate low-dimensional representations, which fails to capture the complex latent relationships between samples.  Moreover, Deep feature screening incorporates a neural network for learning low-dimensional representations and a multivariate rank distance correlation measure (applied on the low-dimensional representations) for feature screening~\cite{li2022deep}. However, the effectiveness of the method needs further investigation. 

Other frequently used feature selection methods include recursive feature elimination~\cite{guyon2002gene} and sequential feature selection~\cite{aha1995comparative}. The former recursively considers smaller and smaller sets of features based on the feature importance obtained by training a classifier. The latter is a greedy algorithm that adds (forward selection) or removes (backward selection) features based on the cross-validation score of a classifier. However, both methods are computational expensive, which become infeasible when dealing with HDLSS data.

% Greedy methods

\section{Method}\label{sec:3}
\begin{figure}[h]
    \centering
    \includegraphics[width=\textwidth]{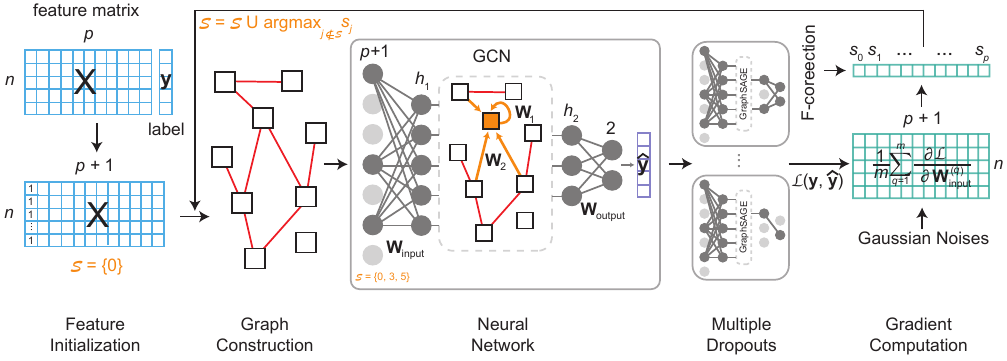}
    \caption{Workflow of GRACES. GRACES consists of feature initialization that adds a bias feature served as the initial selected feature, graph construction with using cosine similarity on the selected features, three-layer neural network with an input linear layer, GraphSAGE layer, and an output linear layer (where gray disks represent the hidden neurons in the neural network), multiple dropouts on the hidden neurons for reducing variance in the subsequent computation, and gradient computation (with introduction of Gaussian noises or F-correction) which gives rise to the current optimal feature according to the gradient magnitude. Note that the activation of the hidden neurons in the input linear layer depends on $\mathcal{S}$.}
    \label{fig:1}
\end{figure}

GRACES is an iterative algorithm which has five major components: feature initialization, graph construction, neural network, multiple dropouts, and gradient computation (Fig. \ref{fig:1}). Motivated by DNP, GRACES aims to iteratively find a set of optimal features which gives rise to the greatest decreases in the optimization loss. For feature initialization, given a feature matrix $\textbf{X}\in\mathbb{R}^{n\times p}$ with $n\ll p$, we first introduce a bias feature (e.g., an all-one column) into \textbf{X} and index it by zero.  The total number of features now is $p+1$, and the original features have the same index numbers as before. We initialize the selected feature set $\mathcal{S}=\{0\}$, i.e., the bias feature. In other words, the bias feature serves as the initial selected feature to start the feature selection process.

For graph construction, we exploit the cosine similarity measure based on the selected features in $\mathcal{S}$. Given two feature vectors $\textbf{x}_i\in\mathbb{R}^{|\mathcal{S}|}$ and $\textbf{x}_j\in\mathbb{R}^{|\mathcal{S}|}$ for sample $i$ and $j$, the cosine similarity is defined as the cosine of the angle between them in the Euclidean space, i.e.,
\begin{equation}
    S_C(\textbf{x}_i,\textbf{x}_j) = \frac{\textbf{x}_i^\top\textbf{x}_j}{\|\textbf{x}_i\|_2\|\textbf{x}_j\|_2}.
\end{equation}
Considering each sample as a node, we connect two nodes if their cosine similarity score is larger than a threshold $\delta$ (which is a hyperparameter of GRACES). The resulting similarity graph captures the latent interactions between samples and will be used in the GCN layer. The similarity graph is different at each iteration, and other similarity measures, such as Pearson correlation and Chi-squared distance \cite{wang2014similarity} (for discrete features), can also be used here. 

We build the neural network with three  layers: an input linear layer, a GCN layer, and an output linear layer. In order to select the features iteratively, we only need to consider weights along the dimensions corresponding to the selected features in the input weight matrix (in other words, for those non-selected features, the corresponding entries in the weight matrix must be zeros) without a bias vector, i.e., 
\begin{equation}
    \hat{\textbf{x}}_j = \text{ReLU}(\textbf{W}_{\text{input}}\textbf{x}_j),
\end{equation}
where $\textbf{x}_j\in\mathbb{R}^{p+1}$ is the feature vector for sample $j$, $\textbf{W}_{\text{input}}\in\mathbb{R}^{h_1\times (p+1)}$ is the learnable weight matrix ($h_1$ denotes the first hidden dimension) such that the $(i+1)$th column is a zero vector for $i\notin \mathcal{S}$.  Subsequently, we utilize one of the classical GCN -- GraphSAGE~\cite{hamilton2017inductive} to refine the embeddings based on the similarity graph constructed from the second step, i.e., 
\begin{equation}
    \tilde{\textbf{x}}_j = \text{ReLU}\Big(\textbf{W}_{\text{1}}\hat{\textbf{x}}_j + \frac{1}{|\mathcal{N}(j)|} \sum_{i\in \mathcal{N}(j)} \textbf{W}_{\text{2}}\hat{\textbf{x}}_i\Big),
\end{equation}
where $\textbf{W}_{\text{1}}\in\mathbb{R}^{h_2\times h_1}$ and $\textbf{W}_{\text{2}}\in\mathbb{R}^{h_2\times h_1}$ are two learnable weight matrices ($h_2$ denotes the second hidden dimension), and $\mathcal{N}(j)$ denotes the neighborhood set of node $j$. GraphSAGE leverages node feature information to efficiently generate embeddings by sampling and aggregating features from a node’s local neighborhood~\cite{hamilton2017inductive}. Finally, the refined embedding is further fed into an output linear layer to produce probabilistic scores of different classes for each sample, i.e., 
\begin{equation}
    \hat{\textbf{y}}_j = \text{Softmax}(\textbf{W}_{\text{output}}\tilde{\textbf{x}}_j + \textbf{b}_{\text{output}}),
\end{equation}
where $\textbf{W}_{\text{output}}\in\mathbb{R}^{h_2\times 2}$ is a learnable weight matrix (assuming the labels are binary, i.e., label zero and label one) and $\textbf{b}_{\text{output}}\in\mathbb{R}^2$ is the bias vector. We  denote the predicted vector containing the probabilities of  label one (second entry in $\hat{\textbf{y}}_j$) for all samples by $\hat{\textbf{y}}\in\mathbb{R}^{n}$.

To reduce the effect of high variance in the subsequent gradient computation, we adopt the same strategy of multiple dropouts as proposed in~\cite{liu2017deep}. After training the neural network based on the selected features, we randomly drop hidden neurons in the GCN layer and the output layer multiple times (the total number of dropouts $m$ is a hyperparameter). In other words, we obtain multiple different dropout neural network models. The technique of multiple dropouts has proved to be effectively stable and robust for deep learning-based feature selection under the HDLSS setting~\cite{liu2017deep, chowdhury2019recurrent}.

For gradient computation, we compute the gradient regarding the input weight for each dropout neural network model and take the mean, i.e., 
\begin{equation}
    \textbf{G} =\frac{1}{m} \sum_{q=1}^m\frac{\partial \mathcal{L} }{\partial \textbf{W}^{(q)}_{\text{input}}} \in\mathbb{R}^{h_1\times (p+1)}
\end{equation}
where $\mathcal{L}$ is the optimization loss, and $\textbf{W}^{(q)}_{\text{input}}$ is the input weight matrix for the $q$th dropout model. Here we use the cross-entropy loss, i.e.,
\begin{equation*}
    \mathcal{L}(\textbf{y}, \hat{\textbf{y}}) = -\frac{1}{n}\sum_{j=1}^n y_j \log{\hat{y}_j} + (1-y_j)\log{(1-\hat{y}_j)},
\end{equation*}
where $y_j$ and  $\hat{y}_j$ are the $j$th entries of $\textbf{y}$ and $\hat{\textbf{y}}$, representing the true label and the predicted probability of label one for sample $j$, respectively. After obtaining the average gradient matrix, the next selected feature can be computed based on the magnitude of the column norm of $\textbf{G}$, i.e., 
\begin{equation}\label{eq:5}
    \mathcal{S} = \mathcal{S} \cup \text{argmax}_{j\notin \mathcal{S}} \|\textbf{g}_j\|_2,
\end{equation}
where $\textbf{g}_j$ is the $j$th column of $\textbf{G}$. The selected feature set is iteratively  updated until reaching the number of requested features, and the final features selected by GRACES is given by $\mathcal{S}$ with the bias feature removed.

To further reduce the effect of overfitting due to low-sample size, we incorporate two additional strategies in GRACES. First, we consider introducing Gaussian noises to the weight matrices of the GCN layer, i.e., adding noise matrices generated from a Gaussian distribution with mean zero and variance $\sigma^2$ (which is a hyperparameter of GRACES) to $\textbf{W}^{(q)}_{\text{1}}$ and $\textbf{W}^{(q)}_{\text{2}}$, for the different dropout models in the gradient computation step. Studies have shown that introduction of Gaussian noises is able to boost the stability and the robustness of deep neural networks during training~\cite{li2020adaptive,jang2021noise,yin2015noisy}. Second, we consider correcting the feature scores (i.e., $\|\textbf{g}_j\|_2$) by incorporating it with the ANOVA F-test, i.e., the final score for feature $j$ is given by
\begin{equation}
    s_j = \alpha g_j + (1-\alpha)f_j,
\end{equation}
where $g_j$ is the normalized score computed from $\|\textbf{g}_j\|_2$, $f_j$ is the normalized score computed from the F-statistic, and $\alpha\in[0,1]$ is the correction weight (which is a hyperparameter of GRACES). Therefore, the  selected feature set is  updated by the follows:
\begin{equation}\label{eq:7}
     \mathcal{S} = \mathcal{S} \cup \text{argmax}_{j\notin \mathcal{S}} s_j.
\end{equation}
The reasons we select the ANOVA F-test are: (1) it is computationally efficient; (2) it achieves a relatively good performance in feature selection for some HDLSS data; (3) it does not suffer from overfitting, so including it can reduce the effect of overfitting in GRACES. The two overfitting-reducing strategies effectively improve the performance of GRACES for HDLSS data.

Detailed steps of GRACES can be found in Algorithm \ref{alg:1}. We  list all the  hyperparameters of GRACES in  Table \ref{tab:0}. Although GRACES is inspired from DNP, it differs from DNP in the following aspects: (1) GRACES constructs a dynamic similarity graph  based on the selected feature at each iteration; (2) GRACES exploits advanced GCN (i.e., GraphSAGE) to refine sample embeddings according to the similarity graph, while DNP only uses MLP which fails to capture latent associations between samples; (3) in addition to multiple dropouts proposed in DNP, GRACES utilizes more overfitting-reducing strategies, including introduction of Gaussian noises and F-correction, to further improve the robustness of feature selection. In the following section, we will see that GRACES significantly outperforms DNP in both synthetic and real-world examples.

\section{Experiments}\label{sec:4}
We evaluated the performance of GRACES on both synthetic and real-world HDLSS datasets along with six representative feature selection methods, including the ANOVA F-test~\cite{st1989analysis}, LR Lasso~\cite{meier2008group}, HSIC Lasso~\cite{yamada2014high}, RF~\cite{breiman2001random}, CancelOut (a traditional deep learning-based feature selection method)~\cite{borisov2019cancelout}, and DNP~\cite{liu2017deep}. HSIC Lasso and DNP are recognized as the state-of-the-art methods for HDLSS feature selection. The reason we chose CancelOut is that it  achieves a relatively better performance compared to other deep learning-based methods (which are not designed specifically for HDLSS data).  We did not compare with GBFS (due to the feature of early stopping), deep feature screening (due to lack of code availability), and recursive feature elimination and sequential feature selection (due to infeasible computation). We used support vector machine (SVM) as the final classifier and the area under the receiver operating characteristic curve (AUROC) as the evaluation metric. All the experiments presented were performed on a Macintosh machine with 32 GB RAM and an Apple M1 Pro chip in Python 3.9. The code of GRACES can be found at \url{https://github.com/canc1993/graces}.

\subsection{Synthetic Datasets}

We used the \texttt{scikit-learn} function \texttt{make\_classification} to generate synthetic data. The function  creates clusters of points normally distributed about vertices of a $q$-dimensional hypercube ($q$ is the number of important features) and assigns an equal number of clusters to each class~\cite{guyon2004result}. We set the number of samples to 60 and fixed the number of important features to 10. We varied the total number of features from 500 to 5000 and considered three synthetic datasets with easy, intermediate, and hard classification difficulty (can be controlled by the variable \texttt{class\_sep}). We randomly split each dataset into 70\% training, 20\% validation, and 10\% testing with 20 replicates. We performed grid search for finding the optimal key hyperparameters for each method. We reported the average test AUROC (over 20 times train test splits) with respect to the total number of features. In the meantime, since we know the exact important features, we also reported the correction rate of the selected features during training. 
% For deep learning-based methods including GRACES, DNP, and CancelOut, we repeatedly ran them for 3 times and took the mean at each train test split. 

\begin{figure}[h]
    \centering
    \includegraphics[width=\textwidth]{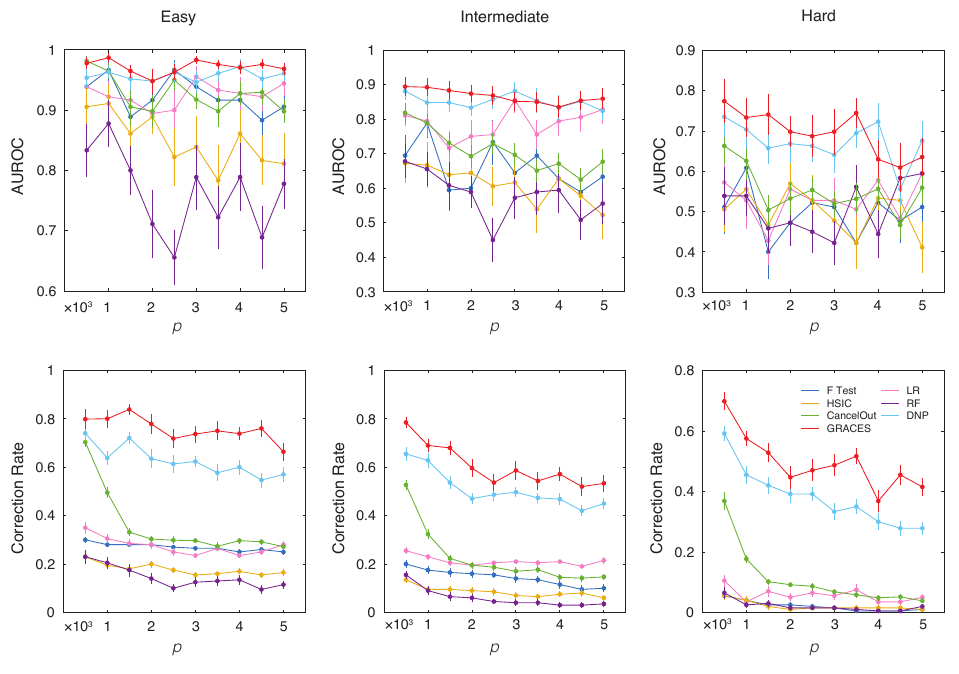}
    \caption{Synthetic datasets. Average test AUROC and correction rate with respect to the total number of features for the easy, intermediate, and hard synthetic datasets. Error bars indicate standard error mean.}
    \label{fig:2}
\end{figure}

The results are shown in Fig. \ref{fig:2}. Clearly, GRACES achieves a superb performance under all three modes. Notably, GRACES is able to capture more correct important features (i.e., the correction rate of GRACES significantly outperforms other methods), which leads to a better test AUROC. Moreover, the performance of GRACES is remarkably stable regarding the increase of the total number of features (especially under the easy and intermediate modes). By contrast, the AUROC of the other methods (except DNP) fluctuates drastically. Under the easy mode, most of the methods (such as the ANOVA F-test, LR Lasso, CancelOut) accomplish a comparable performance (i.e., AUROC > 90\%) even though their correction rates are much lower than that of GRACES. Under the hard mode, however, these methods become ineffective (i.e., AUROC $\sim$ 50\%). Finally, DNP achieves the second-best performance for the three synthetic datasets.

\subsection{Real Datasets}
We used the following six biological datasets:
\begin{itemize}
    \item Colon: Gene expression data from colon tumor patients and normal control;
    \item Leukemia: Gene expression data from acute lymphoblastic leukemia (ALL) patients and normal control;
    \item ALLAML: Gene expression data from acute lymphoblastic leukemia (ALL) patients and acute myeloid leukemia (AML) patients;
    \item GLI\_85: Gene expression data from glioma tumor patients and normal control;
    \item Prostate\_GE: Gene expression data from prostate cancer patients and normal control;
    \item SMK\_CAN\_187: Gene expression data from smokers with lung cancer and smokers without lung cancer.
\end{itemize}
The statistics of the datasets are shown in Table \ref{tab:1}. All the datasets can be downloaded from \url{https://jundongl.github.io/scikit-feature/datasets.html}.

\begin{figure}[h]
    \centering
    \includegraphics[width=\textwidth]{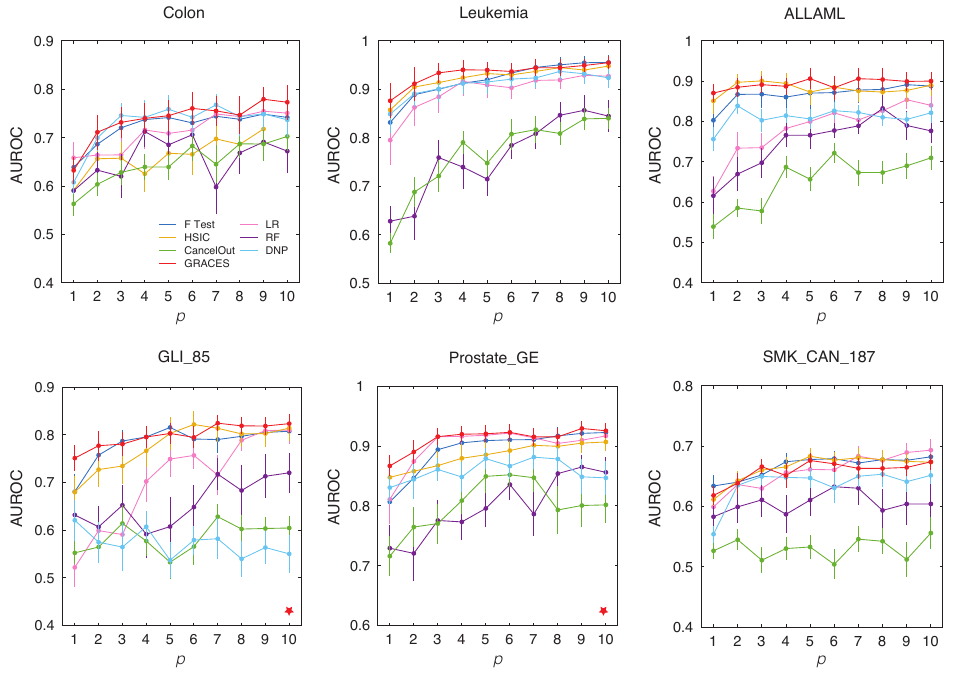}
    \caption{Real-world datasets. Average test AUROC with respect to the number of selected features for each dataset. Error bars indicate standard error mean, and red stars indicate statistical significance compared to the second-best method (p-value < 0.05, one-sample paired t-test on the total 200 data points). }
    \label{fig:3}
\end{figure}

\begin{figure}[h]
    \centering
    \includegraphics[scale=1.8]{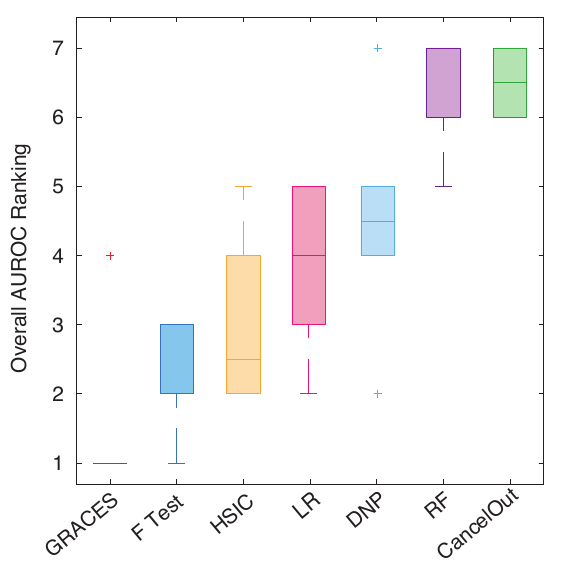}
    \caption{Boxplot of overall AUROC mean ranking over the six datasets for all the feature selection methods.}
    \label{fig:4}
\end{figure}

We randomly split each dataset into 20\% training, 50\% validation, and 30\% testing with 20 replicates. We chose a such low-training size is that a high-training size would result in an extremely high performance for every method (which can be seen in the DNP paper~\cite{liu2017deep}). We performed grid search for finding the optimal key hyperparameters for each method. We reported the average test AUROC (over 20 times train test splits) with respect to the number of selected features from 1 to 10. 
% For deep learning-based methods including GRACES, DNP, and CancelOut, we repeatedly ran them for 3 times and took the mean at each train test split. 
The results are shown in Fig. \ref{fig:3}, where GRACES outperforms the other methods for all the datasets except SMK\_CAN\_187. In particular, on the GLI\_85 and Prostate\_GE datasets, the advantage of GRACES can be shown with statistical significance compared to the second-best method (p-value < 0.05, one-sample paired t-test on the total 200 data points). On the SMK\_CAN\_187 dataset, GRACES still achieves a comparable performance with the well-performing methods. Moreover, the performance of GRACES is stable and robust across all the datasets, while the other methods (such as LR Lasso, HSIC Lasso and DNP) would fail on certain datasets (e.g., LR Lasso on ALLAML; HSIC Lasso on Colon; DNP on GLI\_85), see Fig. \ref{fig:4} and Table \ref{tab:2}. By combining the six dataset, the overall performance of GRACES is significantly better than these of all the other methods (p-value < $10^{-6}$, one-sample paired t-test on the total 1200 data points). Surprisingly, the ANOVA F-test achieves a relative good and stable performance on the real-world datasets. RF and CancelOut, which are not suitable for  HDLSS data, do not perform well. In summary, GRACES can achieve a comparable or improved performance over the baselines on the six biological datasets.

\section{Discussion}\label{sec:5}
Both the synthetic and real-world datasets demonstrate compelling evidence that GRACES can achieve a superb and stable performance on HDLSS datasets. We also computed the total computational time of each method for running the six biological datasets with selected features from 1 to 10, see Table \ref{tab:3}. The ANOVA F-test is the most computationally efficient method among the seven methods. On the other hand, our method GRACES requires more computation resources in finding the optimal features due to its complex architecture. When the number of sample is small (e.g., Colon, Leukemia, ALLAML), the computational time of GRACES is still reasonable. However, when the number of samples becomes large (e.g., SMK\_CAN\_187), the computational time increases drastically. Therefore, GRACES is only suitable for HDLSS data and cannot handle normal feature selection tasks with large-sample sizes.

In this paper, we propose a GRACES model to perform feature selection on HDLSS data. By utilizing GCN along with different overfitting-reducing strategies including multiple dropouts, introduction of Gaussian noises, and F-correction, GRACES achieves a superior performance on both the synthetic and real-world HDLSS datasets compared to other classical feature selection methods. We plan to apply GRACES to more biological datasets that suffer from HDLSS problem, such as different multi-omics data. It will be useful to investigate more sophisticated network architecture to learn the low-dimensional representations of data. For example, hypergraph convolutional network~\cite{feng2019hypergraph,bai2021hypergraph,chen2022survey}, generalized from GCN, is able to exploit  higher-order associations among samples, which might result in a more accurate representation for each sample. Further, more overfitting-reducing techniques such as normalization can be considered.

\newpage
\section*{Figures}
\vfill
\vfill

\newpage
\null
\vfill
\vfill

\newpage
\null
\vfill
\vfill

\newpage
\null
\vfill

\vfill

\newpage
\vfill

\section*{Tables}
\vfill
\begin{table}[h]
\centering
\caption{Hyperparameters of GRACES.}
% \vspace{0.5cm}
\begin{tabularx}{7.5cm}{c *{2}{Y}}
\toprule
Hyperparameter  & Notation  \\
\midrule
Number of Requested Feature & $K$\\
Similarity Score Threshold & $\delta$ \\
First Hidden Dimension & $h_1$ \\
Second Hidden Dimension & $h_2$ \\
Learning Rate & $l$\\
Number of Dropout & $m$ \\
Gaussian Variance & $\sigma^2$ \\
Correction Rate & $\alpha$ \\
\bottomrule
\end{tabularx}
\label{tab:0}
\end{table}
\vfill

\newpage
\null
\vfill
\begin{table}[h]
\centering
\caption{Statistics of the real-world datasets.}
% \vspace{0.5cm}
\begin{tabularx}{17.5cm}{c *{7}{Y}}
\toprule
Dataset  & Colon  & Leukemia &  ALLAML & GLI\_85 & Prost.\_GE & SMK.\_187\\
\midrule
\# Samples & 62 & 72 & 72 & 85& 102& 187\\
\# Features & 2,000 & 7,070 & 7,129 &22,283 & 5,966& 19,993\\
\# Classes & 2 & 2 & 2 & 2 & 2 & 2\\
\bottomrule
\end{tabularx}
\label{tab:1}
\end{table}
\vfill

\newpage
\null
\vfill
\begin{table}[h]
\centering
\caption{Overall AUROC mean and rankings of the feature selection methods. We ranked these methods based on the overall AUROC mean.}
% \vspace{0.5cm}
\begin{tabularx}{17.5cm}{c *{7}{Y}}
\toprule
Dataset  & Colon  & Leukemia &  ALLAML & GLI\_85 & Prost.\_GE & SMK.\_187\\
\midrule
F-Test & 0.7226 (3) & 0.9187 (3) & 0.8674 (3) & 0.7825 (2) & 0.8946 (3) & \textbf{0.6667 (1)} \\
LR Lasso & 0.7124 (4) & 0.8960 (5) & 0.7821 (5) & 0.7040 (4) & 0.9004 (2) & 0.6587 (3)\\
HSIC Lasso &  0.6627 (5) & 0.9226 (2) & 0.8815 (2) & 0.7763 (3) & 0.8845 (4) & 0.6647 (2) \\
RF & 0.6576 (6) & 0.7616 (7) & 0.7477 (6) & 0.6570 (5) & 0.7992 (7) & 0.6057 (6)\\
CancelOut & 0.6478 (7) & 0.7638 (6) & 0.6510 (7) & 0.5843 (6) & 0.8004 (6) & 0.5307 (7)\\
DNP & 0.7291 (2) & 0.9100 (4) & 0.8102 (4) & 0.5716 (7) & 0.8586 (5) & 0.6362 (5)\\
GRACES & \textbf{0.7374 (1)} & \textbf{0.9326 (1)} & \textbf{0.8931 (1)} & \textbf{0.7985 (1)} & \textbf{0.9124 (1)} & 0.6586 (4)\\
\bottomrule
\end{tabularx}
\label{tab:2}
\end{table}
\vfill

\newpage
\null
\vfill
\begin{table}[h]
\centering
\caption{Total computational time with selected features from 1 to 10 (in seconds) of each method for the six biological datasets. We used the default hyperparameters of each method to obtain the computational time.}
% \vspace{0.5cm}
\begin{tabularx}{17.5cm}{c *{7}{Y}}
\toprule
Dataset  & Colon  & Leukemia &  ALLAML & GLI\_85 & Prost.\_GE & SMK.\_187\\
\midrule
F-Test & 0.03 & 0.08 & 0.24 & 0.73 & 0.20 &  1.30 \\
LR Lasso & 0.16 & 0.57 & 0.66 & 1.74 & 0.85 & 3.81\\
HSIC Lasso &  4.19  & 7.42 & 7.48 & 16.58 & 7.59 & 27.64\\
RF & 0.52 & 0.63 & 0.88 & 1.62 & 0.90 & 3.60\\
CancelOut & 1.37 & 5.14 & 5.21 & 12.84 & 5.60 & 26.03\\
DNP & 5.05 & 5.53 & 5.57 & 8.99 & 6.16 & 13.59\\
GRACES & 4.93 & 12.27 &12.02 & 33.66 & 16.29 &128.37\\
\bottomrule
\end{tabularx}
\label{tab:3}
\end{table}
\vfill

% \newpage
% \null
% \vfill
% \begin{table}[h]
% \centering
% \caption{Overall AUROC mean and rankings of the feature selection methods. We ranked these methods based on the overall AUROC mean.}
% \vspace{0.5cm}
% \begin{tabularx}{17.5cm}{c *{7}{Y}}
% \toprule
% Dataset  & Colon  & Leukemia &  ALLAML & GLI\_85 & Prost.\_GE & SMK.\_187\\
% \midrule
% F-Test & 0.7250 (3) & 0.8947 (3) & 0.8496 (3) & \textbf{0.7729 (1)} & 0.8958 (3) & \textbf{0.6602 (1)} \\
% LR Lasso & 0.7143 (4) & 0.8832 (5) & 0.7715 (5) & 0.7379 (4) & 0.8991 (2) & 0.6567 (2)\\
% HSIC Lasso &  - (5) & 0.9006 (2) & 0.8575 (2) & 0.7641 (3) & 0.8669 (5) & 0.6473 (4) \\
% RF & 0.6709 (6) & 0.7661 (6) & 0.7454 (6) & 0.6600 (5) & 0.8174 (7) & 0.6070 (6)\\
% CancelOut & 0.6489 (7) & 0.7522 (7) & 0.6494 (7) & 0.5977 (6) & 0.8249 (6) & 0.5482 (7)\\
% DNP & 0.7252 (2) & 0.8928 (4) & 0.7878 (4) & 0.5810 (7) & 0.8710 (4) & 0.6416 (5)\\
% GRACES & \textbf{0.7377 (1)} & \textbf{0.9113 (1)} & \textbf{0.8709 (1)} & 0.7686 (2) & \textbf{0.9083 (1)} & 0.6567 (2)\\
% \bottomrule
% \end{tabularx}
% \label{tab:2}
% \end{table}
% \vfill

\newpage 
\begin{algorithm}[t]
\caption{GRACES}
\label{alg:1}
\begin{algorithmic}[1]
\STATE{\textbf{Input:} Feature matrix $\textbf{X}\in\mathbb{R}^{n\times p}$, label vector $\textbf{y}\in\mathbb{R}^n$, the number of requested feature $K$, score threshold $\delta$, hidden dimensions $h_1$ and $h_2$, learning rate $l$, number of dropouts $m$, Gaussian variance $\sigma^2$, and correction rate $\alpha$}\\
\STATE{Introduce a bias feature into $\textbf{X}$ and index it by 0}\\
\STATE{Initialize $\mathcal{S}=\{0\}$}\\
 \WHILE{$|\mathcal{S}|\leq K+1$}
\STATE{Construct a cosine similarity graph based on $\mathcal{S}$ with a similarity score threhold $\delta$}\\
\STATE{Train a neural network on $\textbf{X}$ and $\textbf{y}$ with learning rate $l$, including an input layer (with $\textbf{W}_{\text{input}}\in\mathbb{R}^{h_1\times (p+1)}$), a GCN layer (with $\textbf{W}_1, \textbf{W}_2\in\mathbb{R}^{h_2\times h_1}$), and an output layer (with $\textbf{W}_{\text{ouput}}\in\mathbb{R}^{h_2\times 2}$ and $\textbf{b}_{\text{output}}\in\mathbb{R}^2$}\\
\STATE{Dropout $m$ times in the GCN and output layers of the neural network}
\STATE{Introduce Gaussian noises (generated from a Gaussian distribution with mean zero and variance $\sigma^2$) to the GCN layer}\\
\STATE{Compute the average gradient regarding the input weight matrix}\\
\STATE{Correct the feature scores by the ANOVA F-test with correction rate $\alpha$}
\STATE{Update the selected feature set by (\ref{eq:7})}
 \ENDWHILE
 \STATE{Drop the bias feature (i.e., the first element) from $\mathcal{S}$}
\STATE{\textbf{Return:} Selected feature set $\mathcal{S}$.}
\end{algorithmic}
\end{algorithm}
\end{document}